\documentclass[manuscript,screen]{acmart}

\AtBeginDocument{%
  \providecommand\BibTeX{{%
    \normalfont B\kern-0.5em{\scshape i\kern-0.25em b}\kern-0.8em\TeX}}}

\setcopyright{acmcopyright}
\copyrightyear{2022}
\acmYear{2022}
\acmDOI{XXXXXXX.XXXXXXX}

\acmConference[AIPR 2022]{2022 5th International Conference on Artificial Intelligence and Pattern Recognition}{September 23-25, 2022}{Xiamen, China}
\acmPrice{15.00}
\acmISBN{978-1-4503-XXXX-X/18/06}

\newcommand{\vecW}{\mathbf{w}}
\newcommand{\vecV}{\mathbf{v}}
\newcommand{\vecH}{\mathbf{h}}

\usepackage{algpseudocode}
\begin{document}

\title{Transfer learning for ensembles: reducing computation time and keeping the diversity}


\author{Shashkov Ilya}
\affiliation{%
  \institution{MIPT, IITP, Skoltech}
  \city{Dolgoprundy}
  \country{Russian Federation}}
\email{speedcuber17@yandex.ru}

\author{Nikita Balabin}
\affiliation{%
  \institution{Skoltech}
  \country{Russian Federation}}

\author{Evgeny Burnaev}
\affiliation{%
  \institution{Skoltech}
  \country{Russian Federation}}

\author{Alexey Zaytsev}
\affiliation{%
  \institution{Skoltech}
  \country{Russian Federation}}




\begin{abstract}
Transferring a deep neural network trained on one problem to another requires only a small amount of data and little additional computation time. 
The same behaviour holds for ensembles of deep learning models typically superior to a single model.
However, a transfer of deep neural networks ensemble demands relatively high computational expenses.
The probability of overfitting also increases.

Our approach for the transfer learning of ensembles consists of two steps: (a) shifting weights of encoders of all models in the ensemble by a single shift vector and (b) doing a tiny fine-tuning for each individual model afterwards. 
This strategy leads to a speed-up of the training process and gives an opportunity to add models to an ensemble with significantly reduced training time using the shift vector. 

We compare different strategies by computation time, the accuracy of an ensemble, uncertainty estimation and disagreement and conclude that our approach gives competitive results using the same computation complexity in comparison with the traditional approach. Also, our method keeps the ensemble's models' diversity higher.
\end{abstract}





\maketitle

\begin{figure}
  \includegraphics[width=\textwidth]{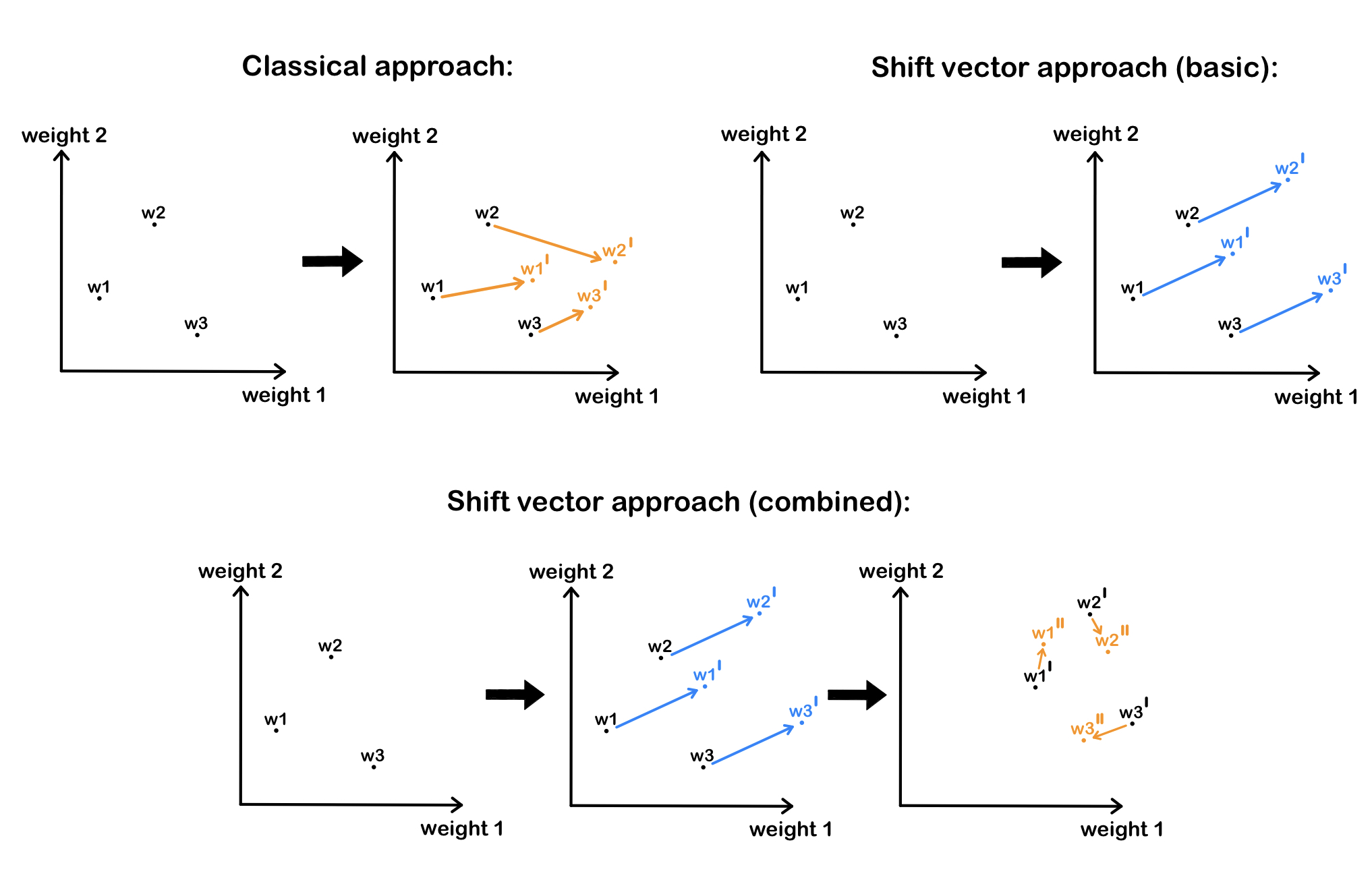}
  \caption{Comparison between considered transfer learning methods for a case of 2-parameter models. \(\vecW_{i}\) is a vector of \(i\)-th model's encoder weights, \(\vecV\) is a shift vector. A classical approach fine-tunes each model in an ensemble separately. a shift approach uses similar shift for all models, limiting the flexibility of fine-tuning, but speeding up the transfer. The combined approach starts with a common shift training an continues with limited number of fine-tuning epoch. Thus, it preserves the robustness and numerical efficiency of the shift approach, while leaving space for more precise fine-tuning.}
  \label{fig:teaser}
\end{figure}

\section{Introduction}

Transfer learning of ensembles is a useful technique that has several benefits. The most significant of them are better predictions compared to any single contributing model and robustness - an ensemble reduces the variance of the predictions and the model performance. Both effects happen mainly to the diversity of single models in the ensemble. However, the training time of an ensemble grows linearly with the number of models in it. Discovering the techniques of reducing the training time of ensembles without affecting models' diversity may greatly facilitate the use of transfer learning of ensembles.

Existing works solve various problems related to ensembles and transfer learning. A paper~\citep{velikanov2022embedded} considers deep learning ensembles that are embedded in a more efficient structure: some parameters of the ensemble's networks are taken from a single reference neural network and shared. The authors identify two regimes for training ensembles: when ensembles train individually and when they share some parts. Another study~\citep{DeepEnsembles} analyses methods of creating ensembles from pre-trained models, keeping them diverse. It suggests a simple algorithm that efficiently identifies a subset of pre-trained models for any downstream dataset. However, it leverages classical methods for training models. Thus, the fine-tuning time is considerable. 

The article~\citep{UncertaintyOverview} shows that uncertainty estimation is an important measure and gives a comprehensive overview of existing methods of evaluating it. It also reviews recent advances in these fields, shows current challenges and gives potential research opportunities. 

The ensemble quality is highly correlated with the diversity \cite{lee2015m, opitz1995generating}. The better the diversity of ensemble the better prediction and uncertainty estimation quality is obtained. The article \cite{garipov2018loss} presented a method for Fast Geometric Ensemble learning, similar to Snapshot ensembling \cite{huang2016snapshot}. The authors obtain fast speed reducing time of learning individual models. Another works \cite{valdenegro2019deep, havasi2020training} proposed another approach to learn ensemble faster - instead of speeding-up invidual ensembles they proposed to learn a one complex model that imitates the real ensemble. However, all these techniques produce ensembles with low diversity and thus with poor uncertainty estimation quality.

In this work, we focus on a shift vector approach to transfer learning and compare it to the classical approach of training every model individually. The shift vector approach is a new method to train models that constitute the ensemble. It enables a single shift of all models with additional fine-tuning afterwards. The ensemble's models are trained jointly, effectively reducing computation time but keeping their diversity high. 

Our approach solves the problem of the efficiency of transfer learning for ensembles in a different way. First, we do not use a reference model but instead use independently trained models initially and shift their weights by a single trained vector. Second, our method does not lose accuracy in the case of using a large number of models. As a result, our models are more diverse in general compared to embedded ensembles. Furthermore, our approach does not compromise the accuracy to lower computational complexity.
From a more practical perspective, the benefits of our method include the ability to train considerable amounts of models to quite a good accuracy without a significant increase in computational time and the ability to use a shift vector to add new models to the ensemble with low or zero time cost. Our research also briefly looks at the uncertainty estimation of deep neural networks' ensembles.
The scheme of our approach is in Figure~\ref{fig:teaser}.

\section{Method}

We consider an ensemble of $n$ models with the same architecture trained for example in an independent way or via single pass~\cite{huang2016snapshot}.
For $i$-th model we have an encoder with parameters $\vecW_i$ and a head $\vecH_i$, $i = 1,\ldots, n$.
Encoder corresponds to producing model embeddings, while aim of the head is to produce decisions on the base on these embeddings.

We separate the model's encoder and head classifier, then train them in different ways. 
Models' classifiers are trained independently, but we aim to find a way to train encoders jointly yet effectively. (see section \textbf{Algorithm for sum loss approach})

We propose the following method for training.
\(\vecW_{i_{0}}\) is the vector of \(i\)-th model's encoder parameters.
The vector \(\vecV\) is a vector of the same dimension as \(\vecW_{i_{0}}\) . 
We initialize it as the mean of \(\vecW_{i_{0}}\) vectors:
\[
    \vecV = \frac1n \sum_{i=1}^{n} \vecW_{i_{0}},
\]
where $n$ is the number of models in the ensemble.
Other options are also possible: it may be filled with zeros, with zeros and ones randomly or taken as a mean of \(\vecW_{i_{0}}\) vectors.

The new weight of models' encoders are:
\[
\vecW_{i_{0, new}} = \vecW_{i_{0}} + \vecV.
\]
So the update of the models in the ensemble share the shift vector \(\vecV\). 
Now we freeze \(\vecW_{i}\) part of new encoders' weights and train only our shift vector. Every iteration after back-propagation models' weights are updated with a new value of \(\vecV\):
\[
    \vecW_{i_{n}} = \vecW_{i_{0}} + \vecV_{n}, 
\]
where \(\vecW_{i_{0}}\) are fixed.

There are several options to train the shift vector $\vecV$. We can:
\begin{enumerate}
    \item  Update the shift vector based on the loss from only one model, chosen randomly on every batch;
    \item Update the shift vector based on  the loss, which is the sum of losses from all ensemble's models;
    \item Update the shift vector based on the total loss from all ensemble's models for a certain amount of epoch and then train models using classical approaches.
\end{enumerate}
Definitely, there exist other options, while they are less important than the idea in general.
In this work, we will investigate all these three options, which vary in terms of required computational resources.

Finally, we have the same update $\vecV$ for all encoders of the ensemble.
Then, for each model we train a separate head classifier freezing an encoder.

\subsection{Algorithm for sum loss approach}

\begin{algorithmic}
\State \(\vecV = \frac1n \sum_{i=1}^{n} \vecW_{i_{0}},\)
\For {\(j \in  [1, ..., num\_epochs]\)}
	\For {\(i \in [1, ..., n]\)}:
		\State Train \(i\)-th model
        \State \(l_{i} \gets\) loss
    \EndFor()
    \State Update classifiers independently using \(l_{i}\)
    \State \(L = \frac{\sum_{i = 1}^{n}l_{i}}{n}\) 
    \State Update \(\vecV\) using \(L\)
    \State \(\vecW_{i} \gets \vecW_{i} + \vecV\)
\EndFor()
\end{algorithmic}

\section{Quality measures}

We compare considered options by ensemble's accuracy, loss and disagreement \cite{kuncheva2003measures} to classical fine-tuning.
Also we are measuring the diversity and the uncertainty estimation of the ensembles. 

As discussed in \cite{fort2019deep}, the lower the accuracy of predictions, the higher its potential mismatch due to the possibility of the wrong answers being random, and then we normalize the Disagreement by the mean accuracy of the predictions.

The disagreement $D_{i,j}$ for a pair of models is the following:
\[
    D_{i,j} = \frac{1}{N (0.5 A_{i} + 0.5 A_{j})} \sum_{n = 1}^N [\hat{y}_i(x_n) \ne \hat{y}_j(x_n)],
\]
where $A_i$ and $A_j$ are accuracies of $i$-th and $j$-th models correspondingly, and $[\cdot]$ is the indicator function we compute for the agreement of predicted labels for objects $x_n$ from the sample of size $N$.
We compute the disagreement of an ensemble as the mean disagreement over all pairs of models in this ensemble.

The uncertainty estimation is done in the following way: we get probabilities for every class of the dataset from a model for every sample and calculate the standard deviations of them for a given ensemble. 
For these uncertainty estimate we decide to reject the object with the uncertainty bigger than the threshold.
After removal, we get the accuracy for the remaining objects. 
Higher accuracies correspond to better uncertainty estimates, as we reject subsample with a big share of erroneous decisions.
In our experiments we use the threshold \(0.065\) for the rejection.

For the experiments with our approach we introduce a relative L2 metric. L2 norm for a vector \(\vecV\) of \(d\) elements may be calculated as $ \| v \|_2 = \sqrt{\sum_{j = 1}^{d} v_{j}^{2}}$.
We calculate a relative L2 metric for the ensemble with $n$ models as:
\[
    L2(\vecV) = \frac{\|\vecV\|_2}{\frac1n \sum_{i=1}^{n} \| \vecW_{i} \|_2}.
\]
This metrics help to see the level of contribution of the shift vector \(\vecV\) to the encoders. It shows that the shift vector has an influence on models' encoders and that encoders are actually modified.

\subsection{Technical details}

To validate our approach, we consider computer vision problems widely used for experiments with deep neural networks.
To train an ensemble we use CIFAR-100 dataset that we transfer to CIFAR-10 dataset~\cite{krizhevsky2009learning}.
The architecture for our model is a ResNet-20~\cite{he2016deep}.

For all experiments, we use the following hyperparameters:
\begin{itemize}
\item $50$ epochs for transfer
\item Cross-entropy loss
\item SGD as an optimizer
\item Learning rate: 1e-1 for a classifier, encoder and shift vector, gradually decreasing to 1e-3
\item Batch size: $128$
\end{itemize}

All computations are done on DataSphere on g1.1 (8 cores, GPU: 1x V100)

\section{Experiments}

In this section, we compare the classical method of transfer learning - fine-tuning with different modifications of our approach with shift vector:
\begin{enumerate}
\item Updating the shift vector based on the loss from only one model, chosen randomly on every batch
\item Updating the shift vector based on the loss, which is the sum of losses from all ensemble's models.
\item Updating the shift vector based on the total loss from all ensemble's models for a certain amount of epoch and then train models using classical 
\end{enumerate}
Also, we will discuss the peculiarities of these methods and how they may be further used.

\subsection{Overall comparison}

\begin{table}[h]
  \caption{Accuracy comparison between all approaches}
      \centering
      \begin{tabular}{lcccccc}
      \hline
    Model & 1& 2& 3&4&5&Ensemble \\
    Method & & & & & & \\
    \hline
      Fine-tuning(50 epochs per model, highest accuracy) & 91.26 \% & 91.05\%& 91.20\%& 91.78\%& 91.40\% & 93.16\% \\
      Fine-tuning(18 epochs per model) & 88.78 \% & 89.08\%& 88.68\%& 89.17\%& 88.92\% & \textbf{90.57\%} \\
     Random model selection (ours) & 80.59\%& 83.83\%& 84.22\%& 80.31\%& 83.65\% & 85.78\% \\
     SUM loss (ours) & 84.55\%& 84.33\%& 85.01\%& 84.91\%& 85.83\% & 88.55\% \\
     SUM loss + Fine-tuning (ours) & 88.85\%& 88.98\%& 88.90\%& 88.93\%& 89.21\% & \textbf{91.01\%} \\
      \hline
      \end{tabular}
\end{table}

There we compare all the approaches. The first row (fine-tuning, 50 epochs per model) shows the highest accuracy that may be reached, but computational complexity is the highest. The next 4 columns have the same computational complexity. From them we conclude that our approach reaches the highest accuracy. Even the classical fine-tuning gives the accuracy that is \(0.44\%\) lower.
Additional details on conducted experiments are presented in Appendix.

\subsection{Uncertainty estimation and disagreement.}

In this section, we are comparing ensembles from previous experiments with the baseline one. 

\paragraph{Disagreement}
Obtained disagreements are in Table~\ref{tab:disagreements}. The ensemble obtained from the experiment with summary loss has the highest disagreement of \(0.168\). 
The lowest has the  ensemble obtained from fine-tuning (\(0.087\)). Our approach gave a disagreement of \(0.108\), which is still higher than the baseline approach gives. High disagreement shows that models are more diverse and therefore the ensemble is more robust. 

\paragraph{Uncertainty estimation}
We also measure the uncertainty estimation. The results are in Table~\ref{tab:disagreements}. There we will use for comparison the fine-tuned ensemble with 18 epochs per model. As expected, we get the highest accuracy improvement for ensemble from sum loss approach. 
The best final accuracy is for our SUM Loss + Fine tuning approach. 
This leads to a conclusion that our approaches with summary loss and with combined training can reach the same accuracy as the traditional method with the same computation complexity.

\begin{table}[h!]
    \centering
    \caption{Disagreement and accuracy values for different approaches for models in an ensemble. Higher disagreement values correspond to more diverse ensembles. We present accuracy in percents before and after rejection of objects with the highest uncertainty, as well as the difference (delta) between these two values. Best values are in bold.}
    \begin{tabular}{ccccc}
    \hline
    & Disagreement & \multicolumn{2}{c}{Accuracy} & Delta \\
    &              & Before rejection & After rejection & accuracy \\
    \hline
    Fine-tuning                       & 0.087 & 90.57 & 93.34 & 2.77 \\
    SUM Loss (ours)                   & $\mathbf{0.168}$ & 88.53 & 93.33 & $\mathbf{4.80}$ \\
    SUM Loss + Fine-tuning (ours)     & 0.108 & $\mathbf{91.01}$ & $\mathbf{93.68}$ & 2.67 \\
    \hline
    \end{tabular}
    \label{tab:disagreements}
\end{table}

\section{Related works}

Here we present articles related to the main aspects of this work. 

\paragraph{Ensembles}
The study ~\citep{ears} uses traditional methods to solve a ear recognition problem using VGG-like networks and building ensembles from these models, getting the best performance with significant improvements over the recently published results on this topic.
Natural ensembling techniques like gradient boosting~\cite{friedman2002stochastic} or gradient boosting with regularization~\cite{kozlovskaia2017deep} don't work well, as single models are too good~\cite{badirli2020gradient}.
Note, that if we use ensembles of binary networks the situation is different~\cite{zhu2019binary}.

\paragraph{Transfer learning}
In the paper~\citep{batteries} 8 DCNN networks were used to solve the problem of capacity estimation of lithium-ion (Li-ion) rechargeable batteries in a low-data regime. These neworks were pre-trained on 10-year daily cycling data from eight implantable Li-ion cells and then transferred to the target task. The performance of the resulting ensemble on the target dataset is compared with that of five other data-driven methods including random forest regression, Gaussian process regression, DCNN, DCNN-TL, and DCNN-EL. The comparison showed that this approach can produce a higher accuracy and robustness than these other data-driven methods in estimating the capacities of the Li-ion cells in the target task.

\paragraph{Diversity-driven ensembles}
Recent study~\citep{Diversity} addresses the diversity and the efficiency of an ensemble by proposing a new method called "Efficient Diversity-Driven Ensemble" (EDDE). This method includes selective transferring of previous generic knowledge for accelerating the training process. The diversity is improved by proposing a new diversity measure and then using it to define a diversity-driven loss function. These techniques are combined by a Boosting-based network. This approach gives the highest accuracy with the lowest training cost compared to other well-known ensemble methods. 
The work ~\citep{Facial} explores the ways to improve accuracy and generalization capability in the field of facial recognition. The proposed Part-based transfer Learning network models how humans recognize facial expressions. It consists of 5 sub-networks, where each one performs transfer learning from one of five subsets of facial landmarks: eyebrows, eyes, nose, mouth, or jaw to expression classification. The resulting model outperforms other approaches and tests show a high generalization capacity, making this approach suitable for real-world usage.

\paragraph{Embedded Ensembles}
Recent paper~\citep{velikanov2022embedded} investigates embedded ensembles both theoretically and empirically. It is a type of ensembles where some parameters of the ensemble's network are shared and taken from a single reference neural network. It was shown that this approach significantly lowers computation complexity and maintains a competitive accuracy.
Another important results that in general deep learning ensembles can be learned via two main regimes: in an independent way~\cite{garipov2018loss,havasi2020training} and in a collective way~\cite{wen2019batchensemble,dusenberry2020efficient}.

\paragraph{Uncertainty estimation}
The paper~\citep{Gal2016UncertaintyID} presented in the field of deep learning some tools to obtain practical uncertainty estimates and developed a theory for them. The second part of this paper links Bayesian modelling and deep learning.

\paragraph{Conclusion}
We see, that our approach is distinguished from other related works by introducing the shift vector.
It reaches slightly higher performance compared to other methods and have similar computation complexity, but a lower training time. 
Moreover, in contrast to some other approaches our shift-based transfer has a large application area and further investigation possibilities, as it doesn't consider a specific problem.

\section{Acknowledgments}

This work by Evgeny Burnaev was supported by the Russian Foundation for Basic Research grant 21-51-12005 NNIO\_a. Other authors were supported by the Russian Foundation for Basic Research grant  20-01-00203.

\section{Conclusions}

Our approach showed competitive results compared to the traditional method of ensemble transferring: we obtain an ensemble of the same accuracy via our combined approach. Uncertainty estimation allows to significantly increase the accuracy of the ensemble obtained from summary loss training. The final accuracy is the same as for the fine-tuned ensemble, but the computation time is lower. The more significant disagreement of ensemble's models in our approaches is an advantage of getting more diverse models with the same accuracy and the same computation time as the traditional fine-tuning. 
The considered relative \(L_{2}\) metric shows that in sum loss and combined approaches the shift vector \(v\) is actually affecting models' encoders and thus the presented method differs from just freezing the encoders.

Also our method gives further opportunities. It may be used in other fields of study and improved from the results of this paper. For example, shift vector \(v\) may be used to significantly reduce training time for new models that have to be added to the ensemble.

Moreover, we can conclude that the optimums on CIFAR-10 and CIFAR-100 datasets have a similar structure: we shifted encoders of 5 models by one vector and still got a high accuracy within a few epochs. Therefore, our method may give better results for tasks of transferring ensembles between datasets with more similar optimums.

\bibliographystyle{ACM-Reference-Format}
\bibliography{references}

\appendix

\section{Additional experiments}

\subsection{Classical method. Fine-tuning}

We will start with a baseline method to have a reference in comparing our methods. Ensemble of 5 ResNet20 models, pre-trained on CIFAR-100, are transferred to CIFAR-10. Every model is trained for 50 epochs individually. 
As expected, we get a gradual increase of every model's accuracy in the learning process. Models achieve an accuracy of around 91 per cent and the ensemble accuracy is 93.16 per cent. This is a pretty common and competitive result that may be used for further comparison. Total training time was around 110 minutes.

\subsection{Shift vector approach. Summary loss}
In this experiment, the loss is calculated as a sum of losses for all ensemble's models, and then back-propagation is done for the shift vector \(v\) based on the resulting loss:
\[
    loss = \sum_{i = 1}^{n}loss_{i}.
\]

The loss is more stable than in the experiment with random model selection. This is an expected result: the shift vector is modified to fit all ensemble’s models better. Accuracy graphs show more gradual growth. Relative L2 is also growing and has a value of about 0.64 at the end of training. This shows that this approach modification is promising because shift vector has a significant impact on models’ encoders. The result is not because of fitting only classifiers with almost unchanged encoders but because of shifting encoders to an optimum for all ensemble’s models.
In this experiment we get the following performance: single models reach an accuracy of about 85 per cent and the ensemble reaches 88.55 per cent. The total training time is about 35 minutes. It is significantly lower than using the fine-tuning approach.

\subsection{Shift vector approach. Combined method}
For the next baseline, we start with the previous approach with summary loss. Then at some point during the training, we switch to the baseline method. Our experiments suggest that the optimal setting is to train models with our approach for 10-15 epochs and then switch to fine-tuning. The shift training continues for 10 epochs before switching; the final fine-tuning goes on for 8 epochs for every model. Such a choice provides a consistent number of epochs for all experiments for our approach. Therefore we get $10 + 5 \cdot 8 = 50$ epochs as in the previous experiments. 

We get around 89 per cent accuracy for a single model and 91.01 per cent accuracy for the ensemble. This result is slightly lower than in the baseline experiment. The total training time was about 45 minutes, which is quite similar to the previous experiment.

Now we will investigate into the computation complexity of this experiment. Theoretically, one epoch in the approach with a summary loss is almost equal to 5 epochs in fine-tuning approach, although, in practice, the computation time is noticeably lower. Let’s take the worst estimate. Let the count unit be 1 run on 1 iteration for 1 model. Let the computation complexity of 1 epoch with summary loss be equal to 5 epochs of fine-tuning. Then we get $(10 \cdot 5 + 8 \cdot 5) \cdot n = 90 \cdot n$ units for this experiment, where n equals to a number of iterations in 1 epoch. This complexity is equal to 90 epochs of fine-tuning (18 epochs per model).
We see that fine-tuning with 18 epochs per model gives the same final accuracy (up to half a per cent). This shows that our approach can reach competitive results (regarding accuracy) compared to classical methods.

\end{document}